\documentclass[final]{IEEEtran}
\ifCLASSINFOpdf
\usepackage[pdftex]{graphicx}
\usepackage{graphicx}
\usepackage{verbatim}
\usepackage{algorithmic}
\usepackage{algorithm2e}
\usepackage{amsmath}
\usepackage{amssymb}

\usepackage{lineno}
\usepackage{amsthm}  
\usepackage{listings}
\usepackage{multirow}
\usepackage{subfigure}
\usepackage{amssymb}
\usepackage{latexsym}
\usepackage{multicol}
\usepackage{epsfig}
\usepackage{float}
\usepackage{xspace}
\usepackage{algorithmwh}
\usepackage{float}

\usepackage[colorlinks=true,linkcolor=blue,citecolor=blue,pdfmenubar=true]{hyperref}

   \graphicspath{{../pdf/}{../jpeg/}}
  \DeclareGraphicsExtensions{.pdf,.jpeg,.png}
\else

\fi

\begin{document}
%

\title{   Towards  prediction  of rapid intensification in   tropical cyclones with recurrent neural networks}
 \author{Rohitash Chandra  
 \thanks{  Dr. Rohitash Chandra is    with Artificial Intelligence and Cybernetics Research Group, Software Foundation, Nausori, Fiji. e-mail  and web:   c.rohitash$@$gmail.com,  http://aicrg.softwarefoundationfiji.org  
 } 
  }

\markboth{   }%
{Shell \MakeLowercase{\textit{et al.}}:  }

\maketitle

\begin{abstract} 

The problem where a  tropical cyclone intensifies dramatically within a short 
period of time is known as rapid intensification. This has been  one of the 
major challenges  for tropical weather forecasting. Recurrent neural networks 
have been promising for time series problems which makes them appropriate for  
rapid intensification.  In this paper,    recurrent neural
networks are used to predict rapid 
intensification cases of tropical cyclones  from   the South Pacific and South 
Indian Ocean regions.  A  class imbalanced problem is encountered  which makes it very challenging to achieve promising 
performance. A simple strategy was proposed to include more positive 
cases for detection where the  false positive rate was slightly improved. The limitations of building an efficient system remains due to the  challenges of addressing the class imbalance problem  encountered for  rapid intensification prediction. This motivates further research in using innovative machine learning methods. 

\end{abstract}


\IEEEpeerreviewmaketitle

\section{Introduction}

   Rapid Intensification (RI) occurs when a  tropical cyclone intensifies 
dramatically within a short period of time \cite{HollidayThompson1979}.  RI 
remains as one of the major challenges in tropical weather forecasting 
\cite{DeMariaEtAl2002,WangZhou2008,cloudcycreview}. This challenge  is partially 
due to  limited understanding of the physical mechanisms in the change in  the 
wind intensity of tropical cyclones  \cite{DeMariaEtAl2002,Gross2001}. It has 
been reported  that warm ocean 
temperatures   and warm-ocean eddies 
 influence the RI of tropical cyclones \cite{ShayEtAl2000,Oropeza2015}. An  
assessment on the relationship between sea surface 
temperature and cyclone wind-intensity revealed no direct relationship in  50 \% 
of the cases studied  \cite{AliCyclone2013}. A comprehensive review has been 
presented  with a focus on RI \cite{cloudcycreview} that considered the 
relationship of different types of cloud formations for cyclogenesis.   Expert 
systems that feature computational intelligence methods have been deployed for 
automatic identification of  weather systems   \cite{Wong2008GACyclone} and 
modelling \cite{Zjavka2016,Wei2012},  hence, they form major motivation for 
tackling  RI.

 Neural networks  are computational intelligence methods that have gained 
recognition in  
application to time series prediction  
\cite{Yanchang2006,Wai-Keung2010,Wai-Keung2010,MeieEmbeddingAntCol2013}. They  
have been involved in cyclone prediction and modelling, 
however, they are  not as popular in weather prediction systems when compared 
with statistical counterparts 
\cite{DeMariaKaplan1994,RozoffKossin2011,Zjavka2016}. 
Although there has not been a comprehensive 
study, the back-propagation neural network has been applied for cyclone track 
prediction \cite{YuanfeiBPNNCyclone2011}. Related  techniques have also been  
used in the area of cyclogenesis  to predict tropical disturbances that 
developed into tropical storms  \cite{Jaiswal2011Cyclogenesis}.

Recurrent neural networks (RNNs) are dynamical systems,   which makes them   
suitable for modelling temporal sequences \cite{Elman_1990}. 
Backpropagation-through-time (BPTT) which employs gradient descent  has been 
widely used for training RNNs 
\cite{Werbos_1990,Bengio_etal1994,frasconi_etal1993}.  

The forecast of the behaviour of cyclones is considered extremely 
important 
for avoiding casualties and mitigating damage to property 
\cite{McBrideHolland1987,Holland2009}. Cyclones behave differently in different 
ocean basins, hence meteorological offices around the world adapt to a 
combination of techniques to predict several interrelated features that include 
tracks, intensity,  and accompanying rainfall  
\cite{McBrideHolland1987,RoyKovordanyi2012}. There has been a number of cyclone track prediction methods and models 
developed for various ocean basins \cite{RoyKovordanyi2012}. 
 Coevolutionary 
RNNs have been applied for cyclone track and wind intensity prediction problem 
for the South Pacific Ocean  with promising results  
\cite{ChandraDayalCEC2015,ChandraDayalIJCNN2015}, which  motivates their 
application to the problem of RI.

 There has been devastating impact  in the countries that fall in the path of 
rapidly intensifying tropical cyclones. There has not been much work done   to 
predict  RI cases, which could be very useful in cyclone  disaster management 
systems.  A study of the factors that cause rapid  intensification  is also very 
important in order to make a robust prediction. Recently, RI was approached for modelling using coevolutionary recurrent neural networks on assumption that RI cases have been previously detected \cite{ChandraRI2015}.Modelling RI refers to the ability of the prediction model to precisely give the value for intensification, i.e. by how many knots will the cyclone intensify in next 24 hours.  Although there exists a number of challenges for modelling, however, detection is the most challenging as the number of cases are very small when compared to the entire dataset of cyclones that occurred in the particular region. This issue will be highlighted in this study  and the challenges of detection will be given.

 In this paper,   an assessment is presented for the duration  of the  
cyclones, the 
number of RI cases and relationship of the duration with the number of RI cases 
 for   the South Pacific and the South Indian 
Ocean regions.   Recurrent neural networks 
are trained using 
back-propagation-through-time \cite{Werbos_1990}  to predict  the occurrence of RI.  
 
 The rest of the paper is organised as follows.   Section 2 gives a background 
of cyclones  and RNNs. In Section 3, the proposed framework is presented, 
while Section 4 presents the  experiments and results. Section 5 concludes the 
paper with the discussion of future work.

\section{  Mythology: Recurrent Neural Networks for  Rapid Intensification}

 \subsection{Background  }

Neural networks used for time series prediction are mainly 
characterized into \textit{feedforward} and \textit{recurrent } architectures
 \cite{Haykin_2008}. Feedforward   networks have one or more hidden layers that propagate information to the output layer. In contrast to feedforward networks, RNNs are dynamical systems whose next state and output depends on the current network state and
input; this makes them   suitable for modelling temporal sequences.  The  Elman   RNN  is  a popular  and efficient   architecture that employs  context units to store the output of the state neurons from the computation of the previous time steps \cite{Elman_1990}. Each layer contains  neurons that propagate activation   from one layer to the next by computing a transfer function of their weighted sum of inputs. The context layer is used for computation of present states as they contain information about the previous states as shown in  Figure  \ref{rnn:speed} which is designed for the RI problem.   The goal of the context layer is a mechanism to  transfer  information from the previous time step to the future   step when presented with data from  the present time step.  The   dynamics of the change of hidden state neuron activation's in Elman RNN is given by Equation (\ref{elmaneq}).

 \begin{equation}
\label{elmaneq}
y_i (t) = f \left( \sum_{k=1}^K{ v_{ik} \; y_k ( t - 1)} + \sum_{j=1}^J{w_{ij} \; x_j (t -1)}\right)  
\end{equation} 
               
\noindent where $y_k (t )$ and $x_j (t)$ represent the output of the context state  neuron and input neurons, respectively. $v_{ik}$ and $w_{ij}$ represent their corresponding weights; $i$ represents the number of input neurons while $j$ and $k$ represents the number of hidden  and context layer neurons, respectively. $f(.)$ is a sigmoid 
transfer function. Time $(t)$ refers to each data point of the time series sample used for prediction.

 Neural networks learn by training on 
data using an algorithm that modifies the  weights as directed
by a learning objective for a particular application. The dataset is usually   divided into   a \textit{training set}  and a \textit{testing set}.  The goal of learning  is to  find the set of weights of the
neural network  on the  given  training data in order to achieve maximum performance on unseen data.  This is done by adjusting the weights in the network
according to a learning rule until a certain criterion is met, which is usually
expressed in terms of the \textit{network output error} or  \textit{cost
function}.   

\subsection{Proposed methodology}

Rapid intensification is defined by an increase in wind intensity by 30 knots 
within 24 hours  \cite{KaplanDeMaria2003}. A system is needed that can detect 
and  predict the occurrence of RI in the next 24 hours.  
We use the  RNN to predict the occurrence of 
RI from past cases. This    can be viewed as a boolean classification 
problem 
where a decision is made (positive  or negative).

\begin{figure*}[htp]
\centerline{\includegraphics[height=46mm]{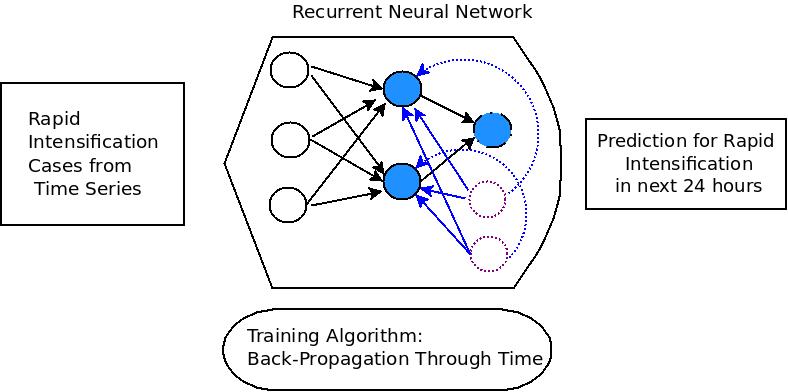}}
\caption{ Backpropagation through-time  used for 
training Elman RNN on the RI cases from the time series data.    }
\label{rnn:overview}
\end{figure*}

 An overview of the  RNN  with 
the 
training data and the respective training algorithms is given in Figure 
\ref{rnn:overview}.


The cyclones used in this study are taken from the South Pacific and South 
Indian Ocean region \cite{jtwc}. The dataset  contains times series information 
about the  wind intensity, month,  and  track information in terms of  
longitude 
and latitude.  The  wind intensity for 
each cyclone is first analysed by a simple rule where a change within 24 hours 
is monitored for every point in the time series. If the wind intensity is 
greater than or equal to  30 knots within  24 hours, then the case is marked as 
positive, otherwise, it is negative.     Note that each data point in the time 
series represents cyclone behavior taken at every six hours for  regular 
intervals.  Therefore, 30 hours of data is used for prediction.  

  The detection of RI is implemented using the following strategies. In 
Strategy I, all the 
positive cases are captured  with RI greater than or equal to  30 knots and 
negative otherwise.     Strategy II considers positive cases with  RI greater 
than or equal to  10 knots and negative otherwise.

   Table \ref{tab:occurance} shows that the number of positive 
 cases of RI is a small portion when compared to negative cases, hence this is 
 a class imbalance problem \cite{Japkowicz2002}. 

\begin{table}[htbp]

\begin{center}
 
\caption{ Detection  of Rapid Intensification   }

\begin{tabular}{ l   l l    l  l }  	  \hline
 Region &Dataset    & No. Positive & No. Negative& Total\\
 	\hline	
   	\hline	
   South Pacific	&Training Set  &     155  &  4798    &  4953\\
 &Testing Set   &       7    & 2002     & 2009  \\

 \hline
 South Indian  &Training Set  &     190  &  6887   &  7077\\
  &Testing Set   &       70    & 6676     & 6746  \\

 \hline
 
 \hline
\end{tabular}
\label{tab:occurance}
\end{center}
 
\end{table}

 RNNs can 
be trained with the principle of the delta learning rule. The general idea 
behind the delta learning rule is to use gradient descent to search the 
hypothesis space of the weight vectors and find the weights that best fit  the 
set of training examples. Gradient descent is one of the most widely
used RNN training in the  implementation as  
backpropagation-through-time (BPTT) \cite{Werbos_1990}.

\begin{algorithm}
\caption{BPTT for Training  Elman RNNs}

\begin{algorithmic}

\small 
\STATE   Initialise the RNN weights with small random numbers in 
range [-0.5, 0.5]\\

\FOR{  each \textit{Epoch} until termination }
   \FOR{ each Sample}   
     \FOR{  $n$ Time-Steps}
\STATE   Forward Propagate
      \ENDFOR
      \FOR{  $n$ Time-Steps}
\STATE i) Backpropagate Errors using Gradient Descent
\STATE ii) Weight update 
 \ENDFOR

 \ENDFOR
 \ENDFOR
 
\label{alg:bptt}
\end{algorithmic}
\end{algorithm}

 The BPTT algorithm unfolds 
a recurrent neural network in time into a deep multilayer feedforward network 
and employs the error backpropagation for weight update as shown in Algorithm 
\ref{alg:bptt}.   When unfolded in time, the network has the same behaviour as 
a recurrent neural network for a finite number of time steps.

\section{Analyses, Experiments and Results}

This section presents  analyses, experiments  and results using   RNNs for  prediction    RI  cases in tropical cyclones.   The  \textit{Smart Bilo: computational intelligence 
framework} implementation of   BPTT for RNN is used for the  
respective 
experiments \cite{smartbilo}.

\subsection{Cyclone Dataset Analyses }
  
  The Southern Hemisphere tropical cyclone best-track data from Joint Typhoon 
Warning Center  recorded every 6-hours is  used as the main source of data 
\cite{jtwc}. We consider only the austral summer tropical cyclone season 
(November to April) from 1980 to 2013 data  in the current study as data prior 
to the satellite era is  not reliable due to inconsistencies and missing 
values. 
The South Indian basin domain is taken to be from 0-30$^{\circ}$S, 
30$^{\circ}$E-130$^{\circ}$E and South Pacific domain is from 0-30$^{\circ}$S, 
130$^{\circ}$E-130$^{\circ}$W. 
  
The original data of tropical cyclone wind intensity in the South Pacific 
\cite{jtwc}  was divided into training  and testing set as follows: 
 
 \begin{itemize}
 \item Training Set: Cyclones from 1985 - 2005 (219 Cyclones) 
 \item Testing  Set: Cyclones from 2006 - 2013  (71 Cyclones)
  \end{itemize}
  
In the case for South Indian Ocean \cite{jtwc}, the details are  as follows: 
 
 \begin{itemize}
 \item Training Set: Cyclones from 1985 - 2001 (  285 Cyclones) 
 \item Testing  Set: Cyclones from 2002 - 2013  (  190 Cyclones )
  \end{itemize}
  
\begin{figure*}[htbp]
  \begin{center}
    \begin{tabular}{c}   
     
    \subfigure[ Training  dataset ]{ \includegraphics[width=2.4in, height=5in , 
angle= 270]{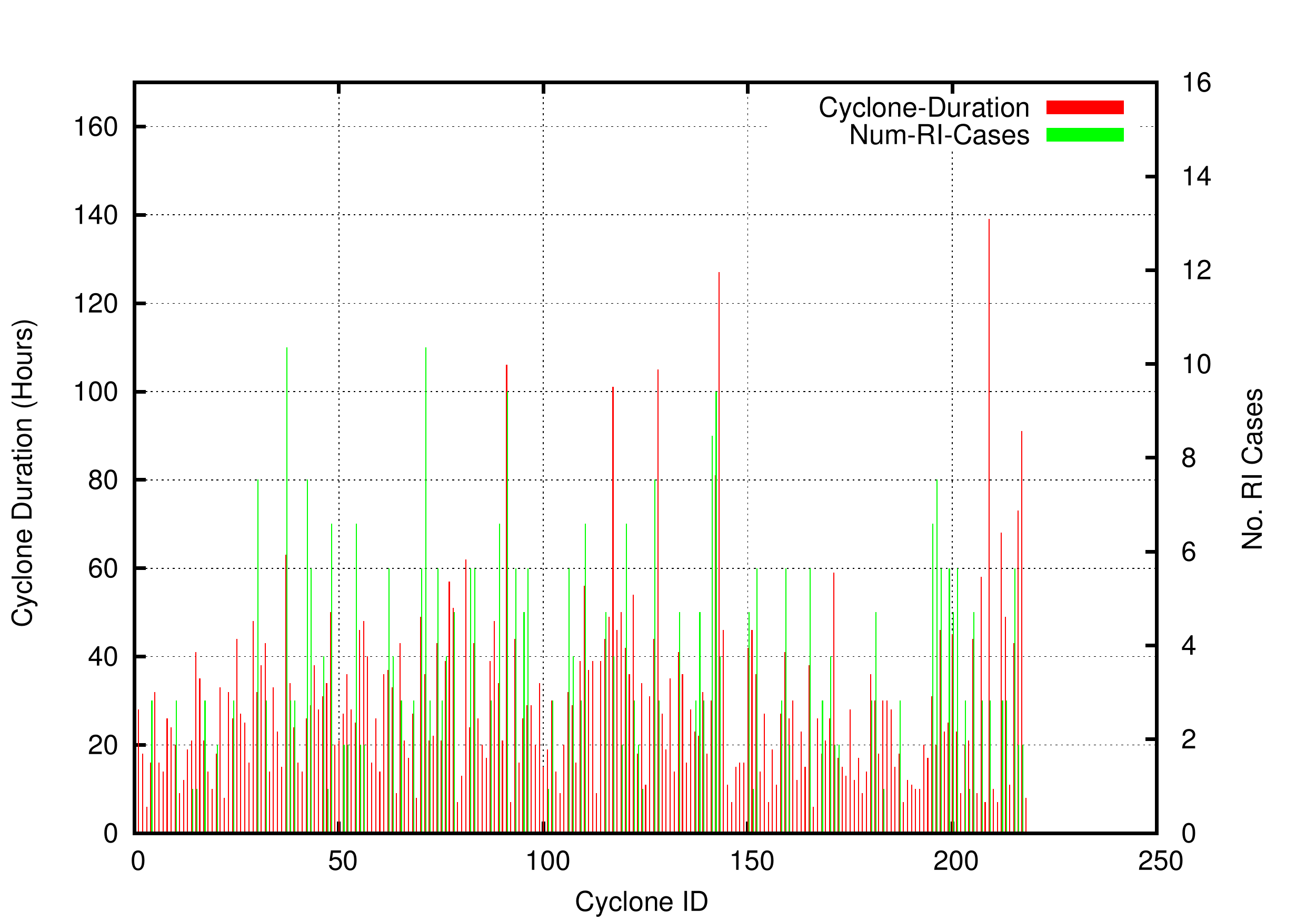}} \\
     \subfigure[  Testing  dataset]{ \includegraphics[width=2.4in, height=5in , 
angle= 270]{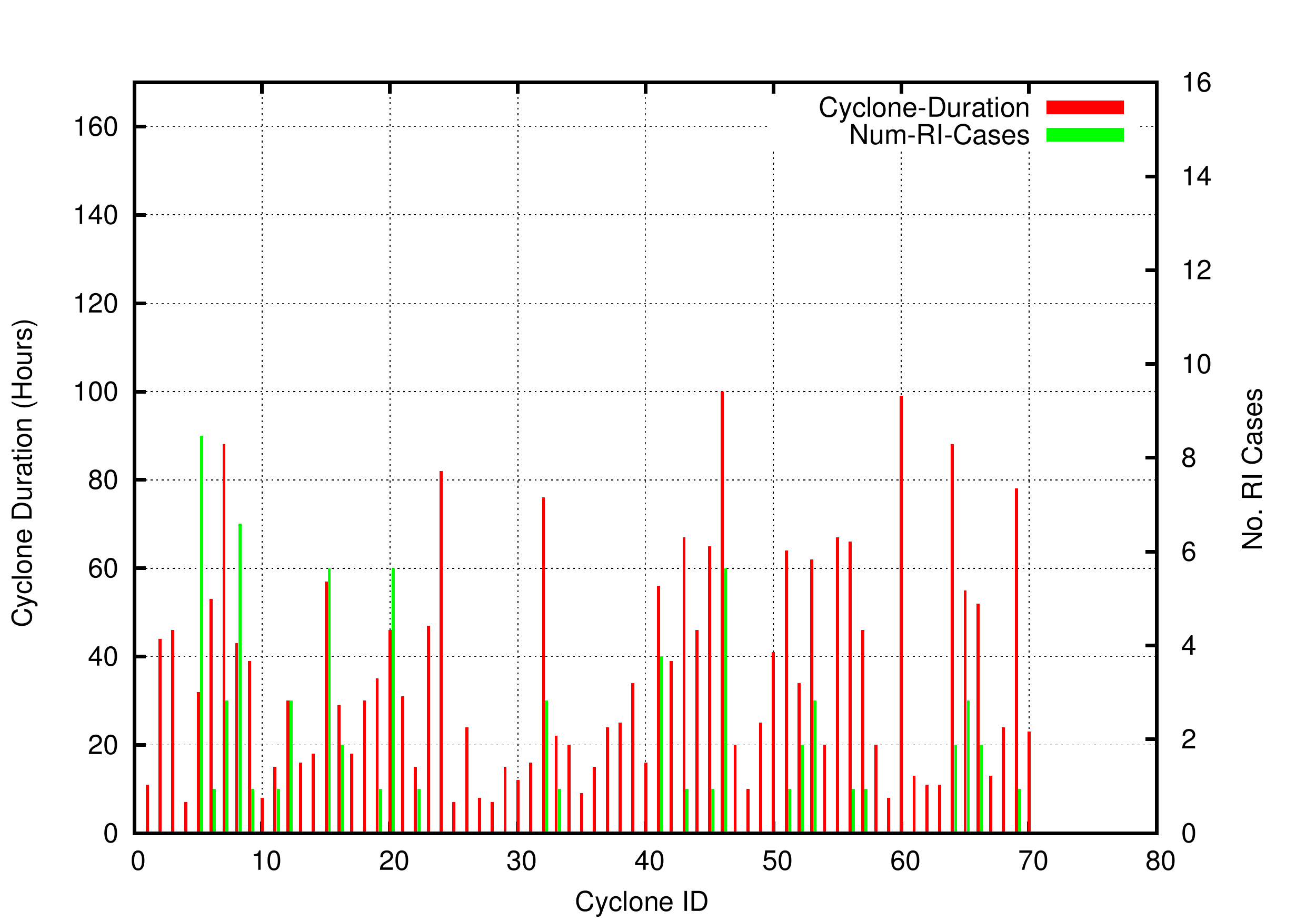}} \\
    \end{tabular}
    \caption{   South Pacific: Number of  RI cases    and duration of each 
cyclone over the cyclone identification number (ID). Each point of cyclone 
duration in y-axis represents 6 hours. Note that for  certain cyclones, there 
are no cases of RI.   }
 \label{fig:CySP}
  \end{center}
\end{figure*}

\begin{figure*}[htbp]
  \begin{center}
    \begin{tabular}{c}   
     
    \subfigure[ Training  dataset ]{ \includegraphics[width=2.4in, 
height=5.8in, angle= 270]{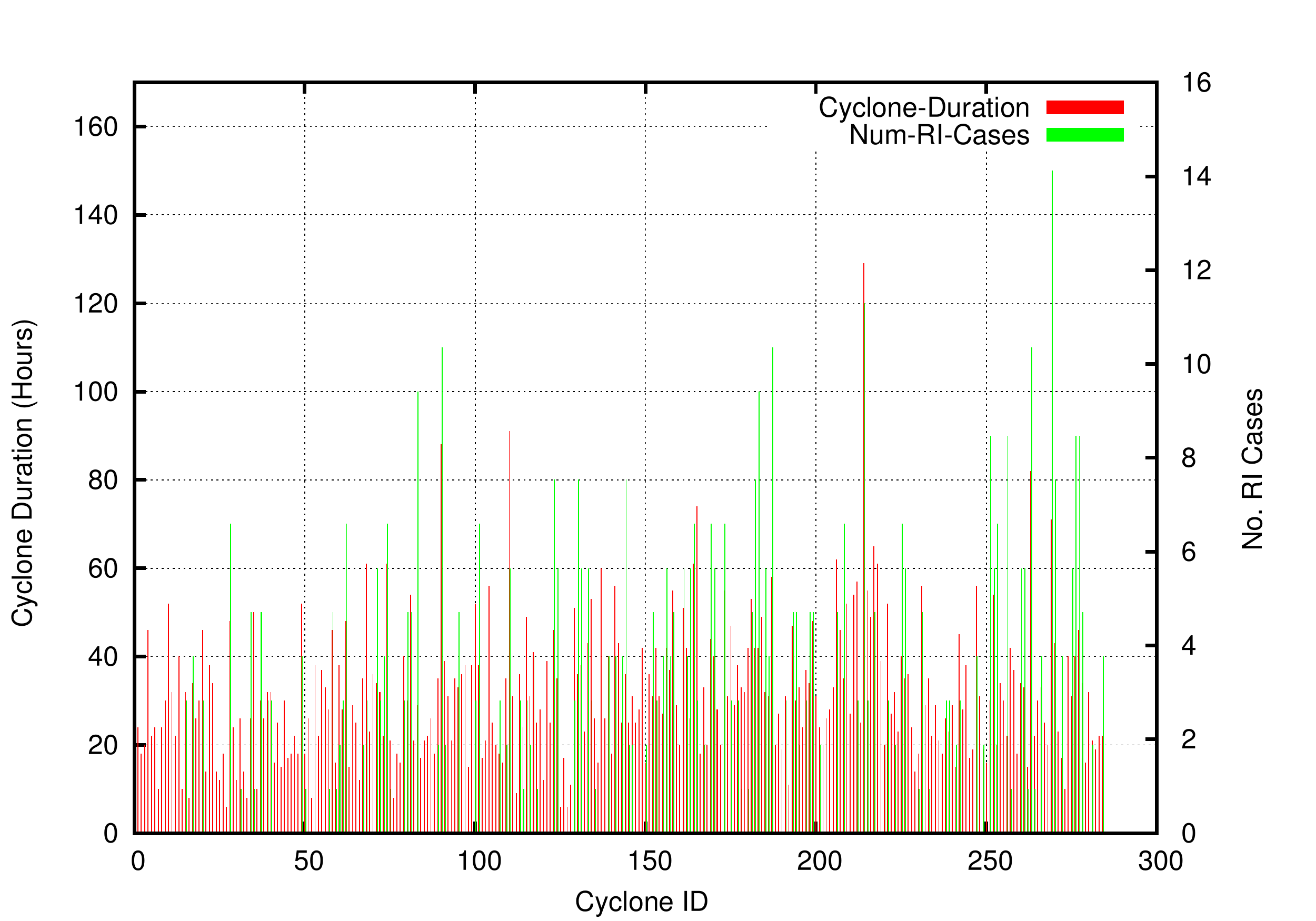}} \\
     \subfigure[Testing  dataset]{ \includegraphics[width=2.4in, height=5.8in , 
angle= 270]{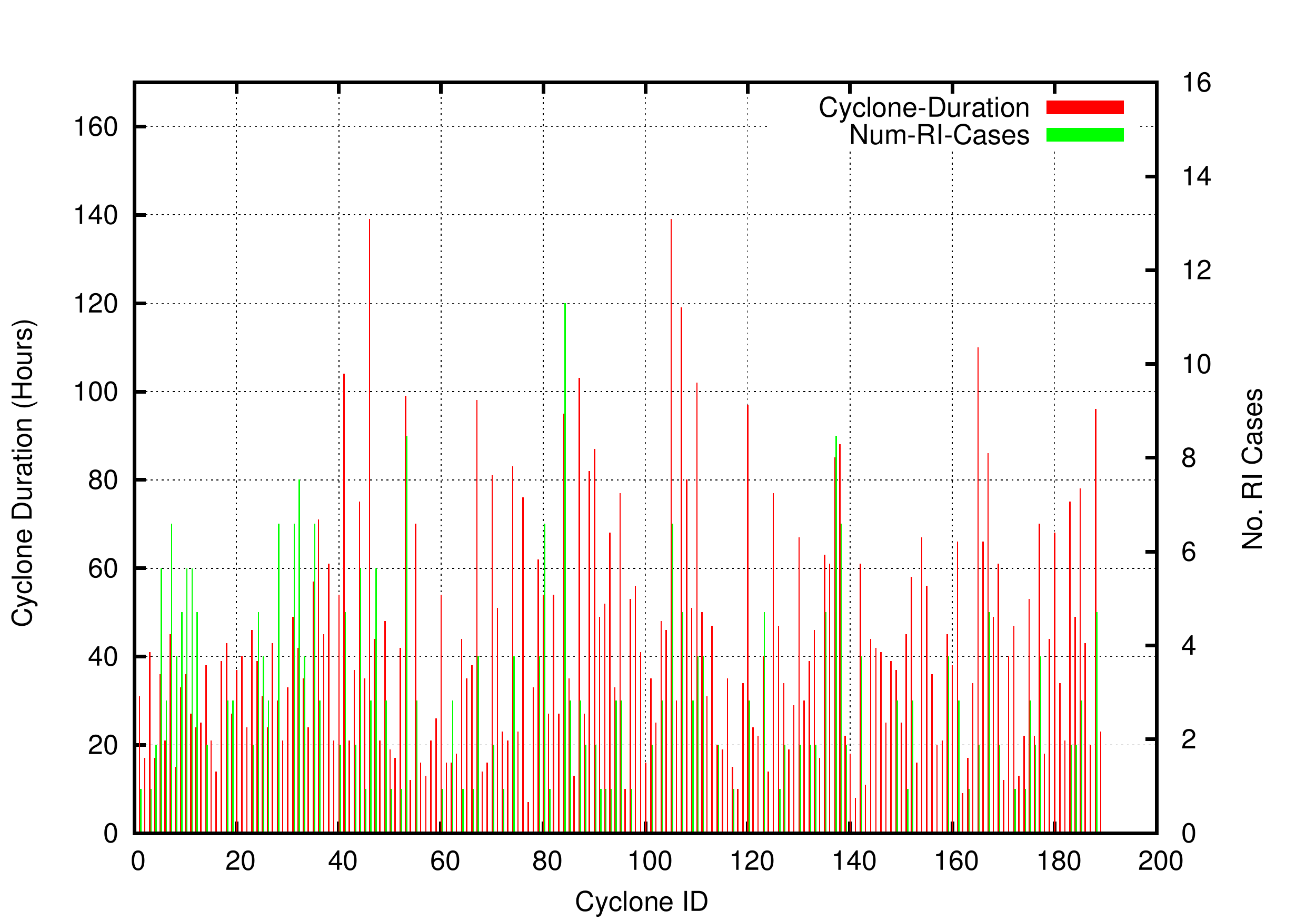}} \\
    \end{tabular}
    \caption{    South Indian Ocean: Number of  RI cases    and duration of 
each 
cyclone over the cyclone identification number (ID). Each point of cyclone 
duration in y-axis represents 6 hours.   Note that for  certain cyclones, there 
are no cases of RI.   }
 \label{fig:CyIO}
  \end{center}
\end{figure*}

 
 Figures \ref{fig:CySP} and \ref{fig:CyIO} show  the details of the duration of 
each cyclone in the training and test dataset for different cyclones given by 
their identification number (ID) on the x-axis.   Note that each point of 
duration in the y-axis  represents 6 hours. The second y-axis  in the histogram 
 
shows the number of cases of RI for each of the corresponding cyclones.    
These 
figures show the relationship between the number of cases of RI   with their 
duration  given by a number of  hours.   It is observed that  in several cases, 
 
the number of cases of RI does not directly relate to the duration of the 
cyclone.  For instance, as shown in Figure 6 (b), Cyclone ID 20 has cyclone 
duration of about 50 $\times$ 6 hours.  It contains  about 6 cases of 
intensification whereas Cyclone ID 5  has a duration of 30 $\times$ 6 hours, 
which contains about 9 cases.

The number of RI cases in the South Indian Ocean is more than the   South 
Pacific for both the training and testing dataset. This is due to the fact that 
the  number of cyclones in the South Indian Ocean is more than  the South 
Pacific.

\subsection{Detection of  Rapid Intensification }

The detection  problem can be viewed as a classification problem 
that involves time series as input and a decision by the RNN whether  it is a 
case of  RI.  The RNN is defined with the topology where 1 
neuron is used in the input layer and 1 in the output layer. The RNN unfolds 
for 
$k$ time steps which are fixed at 5, this  corresponds to the  data  points 
captured in 30 hours.  

\begin{figure*}[htb]
  \begin{center}
    \begin{tabular}{cc} 
      \subfigure[ROC South Pacific Ocean]{\includegraphics[width=60mm, angle= 
270]{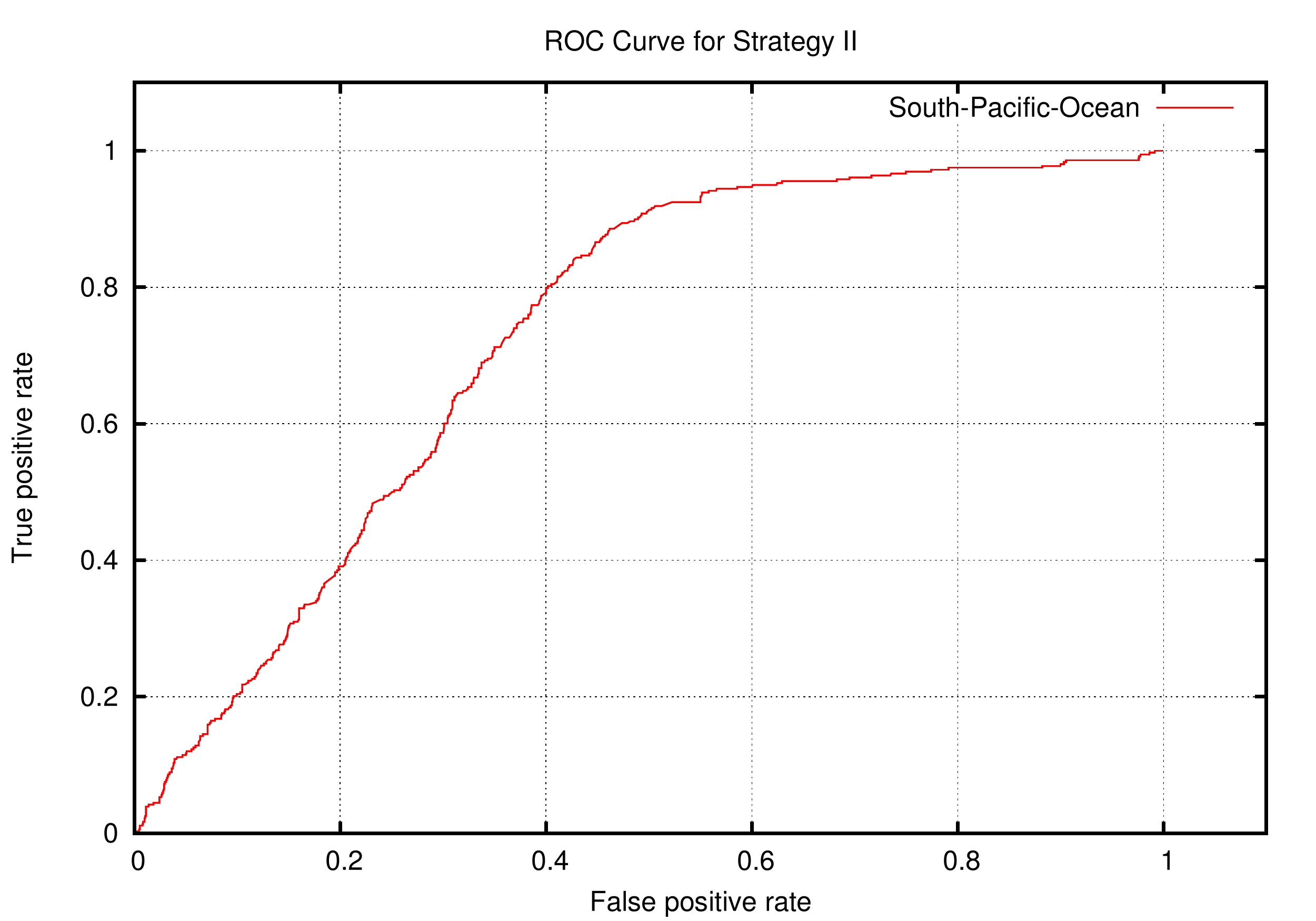}}
    \subfigure[ROC South Indian Ocean]{\includegraphics[width=60mm, angle= 
270]{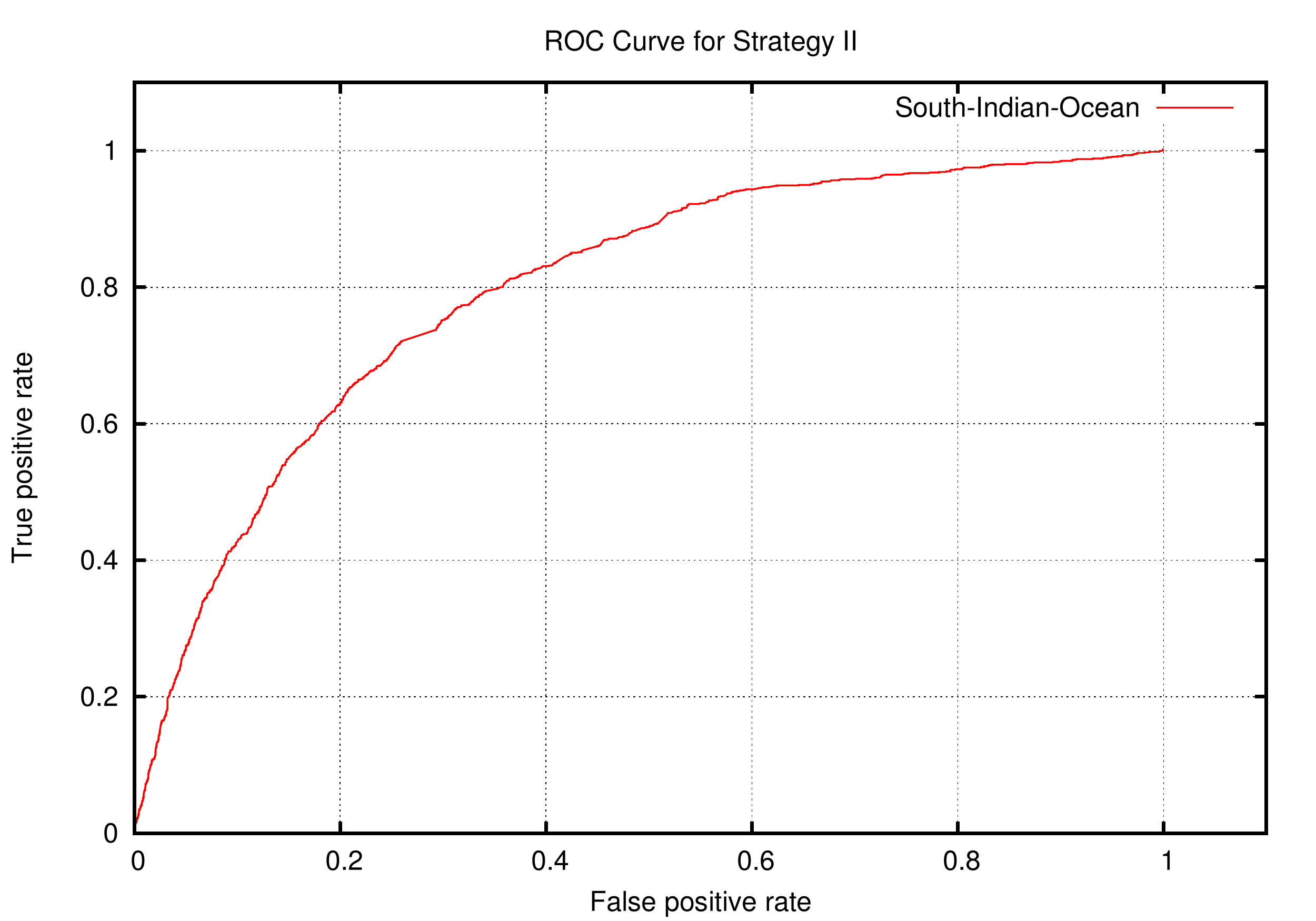}} \\
    \end{tabular}
    \caption{Receiver operating characteristic (ROC) curves  for detection of rapid intensification cases in Strategy II for South  Pacific and South Indian Ocean.  }
 \label{fig:roc}
  \end{center}
\end{figure*}

The extracted dataset is composed of  positive and negative  cases of RI in the 
training and test set as shown in Table \ref{tab:occurance}. The results for  
the detection  (positive and negative cases) of RI in the South 
Pacific and South Indian Ocean region are given in Table \ref{tab:classify-IO}. 
Note that the results are given for two different strategies (Strategy I and 
II) which use 5 and 10 hidden neurons, respectively. In Strategy I, the 
distinction between positive and negative cases are made when rapid 
intensification is by 30 knows. This results in a highly unbalanced datasets 
which makes detection of true positives very difficult. Hence, Strategy II is 
used where the distinction between positive and negative cases is when rapid 
intensification is by 10 knots. In Strategy II, we achieve a bit 
poorer generalisation 
performance when compared to Strategy I, however, there is better detection of 
rapid intensification cases as shown by rate of true positives in Tables 
\ref{tab:confuSP} - \ref{tab:St2confuIO}. These tables report  the 
best result from the 
30 experimental runs. The  receiver operating characteristic (ROC) further 
describes the behaviour of the RNN detection system for Strategy II from 
best experimental run given in 
Figure \ref{fig:roc}. We note that the results show that the RI problem is very 
challenging and there is a need to improve the performance for detection of true 
positives  with innovative  strategies in learning unbalanced data sets.

\begin{table}[htbp]

\begin{center}
 \small 
\caption{ Detection of  RI    }

\begin{tabular}{   c c    c   c }  	  \hline

  Problem & Strategy    & Percentage   (Test)  \\	
							 	
   	\hline	  
 South Indian & I  &    97.390  $ \pm $  0.008   \\  
 
 South Indian & II  &   81.736   $ \pm $ 0.219      \\ 
 \hline 
 South Pacific & I   &  97.214 $ \pm $  0.013 \\  
  South Pacific & II   &    79.779 $ \pm $  0.169      \\  
 \hline
\end{tabular}
\label{tab:classify-IO}
\end{center}
 
\end{table}

\begin{table}[htbp]
\begin{center}
 \small 
\caption{ Strategy  I Confusion Matrix for South Pacific     }
\begin{tabular}{l|l|c|c|c}
\multicolumn{2}{c}{}&\multicolumn{2}{c}{Predicted}&\\
\cline{3-4}
\multicolumn{2}{c|}{}&Positive&Negative&\multicolumn{1}{c}{Total}\\
\cline{2-4}
\multirow{2}{*}{Actual}& Positive & $0$ & $7$ & $7$\\
\cline{2-4}
& Negative & $2$ & $1999$ & $2001$\\
\cline{2-4} 

\multicolumn{1}{c}{} & \multicolumn{1}{c}{Total} & \multicolumn{1}{c}{$2$} & 
\multicolumn{1}{c}{$2006$} & \multicolumn{1}{c}{$2008$  }\\

\end{tabular}	
\label{tab:confuSP}
\end{center}
\end{table}

\begin{table}[htbp]
\begin{center}
 \small 
\caption{ Strategy I Confusion Matrix for Indian Ocean     }
\begin{tabular}{l|l|c|c|c}
\multicolumn{2}{c}{}&\multicolumn{2}{c}{Predicted}&\\
\cline{3-4}
\multicolumn{2}{c|}{}&Positive&Negative&\multicolumn{1}{c}{Total}\\
\cline{2-4}
\multirow{2}{*}{Actual}& Positive & $0$ & $70$ & $70$\\
\cline{2-4}
& Negative & $7$ & $6669$ & $6676$\\
\cline{2-4}

\multicolumn{1}{c}{} & \multicolumn{1}{c}{Total} & \multicolumn{1}{c}{$7$} & 
\multicolumn{1}{c}{$ 6739$} & \multicolumn{1}{c}{$6746 $}\\

\end{tabular}	
\label{tab:confuIO}
\end{center}
\end{table}
  

\begin{table}[htbp]
\begin{center}
 \small 
\caption{ Strategy  II Confusion Matrix for South Pacific     }
\begin{tabular}{l|l|c|c|c}
\multicolumn{2}{c}{}&\multicolumn{2}{c}{Predicted}&\\
\cline{3-4}
\multicolumn{2}{c|}{}&Positive&Negative&\multicolumn{1}{c}{Total}\\
\cline{2-4}
\multirow{2}{*}{Actual}& Positive & $50$ & $308$ & $358$\\
\cline{2-4}
& Negative & $66$ & $1452$ & $1518$\\
\cline{2-4}

\multicolumn{1}{c}{} & \multicolumn{1}{c}{Total} & \multicolumn{1}{c}{$116$} & 
\multicolumn{1}{c}{$1760$} & \multicolumn{1}{c}{ 1876}\\

\end{tabular}	
\label{tab:St2confuSP}
\end{center}
\end{table}

\begin{table}[htbp]
\begin{center}
 \small 
\caption{ Strategy II Confusion Matrix for Indian Ocean     }
\begin{tabular}{l|l|c|c|c}
\multicolumn{2}{c}{}&\multicolumn{2}{c}{Predicted}&\\
\cline{3-4}
\multicolumn{2}{c|}{}&Positive&Negative&\multicolumn{1}{c}{Total}\\
\cline{2-4}
\multirow{2}{*}{Actual}& Positive & $381$ & $837$ & $1218 $\\
\cline{2-4}
& Negative & $316 $ & $4835$ & $5151$\\
\cline{2-4}

\multicolumn{1}{c}{} & \multicolumn{1}{c}{Total} & \multicolumn{1}{c}{$697$} & 
\multicolumn{1}{c}{$ 5672$} & \multicolumn{1}{c}{ 6369}\\

\end{tabular}	
\label{tab:St2confuIO}
\end{center}
\end{table}

\subsection{Discussion}

 After extraction of the cases for RI, 
it was determined that  the detection  problem features a  class imbalanced 
problem that featured more than  98 \% negative cases 
when RI of 30 knots was considered as Strategy I. Hence, Strategy II 
gathered  more positive cases   where RI of more than 10 knots 
was considered as positive. This lead to the RNN with more true positive cases 
when compared to Strategy I. However, the quality  of detection becomes an 
issue when the difference between negative and positive cases is lowered from 30 
to 10 knots.

Furthermore, BPTT has shown 
that the track of  the cyclone is an important factor  for modelling   RI.  The 
track information makes modelling  RI a  multi-variate time series problem. The 
comparison of the RNN performance with some of the prominent backpropagation 
algorithms from the literature shows that it is promising  for the modelling 
phase.     The 
experiments only 
considered 5 data points which are taken every 6 hours and spans for 30 hours. 
Better convergence could be obtained when more data points are available, i.e 
if cyclone readings are taken every  2 or 3 hours. Therefore, the proposed 
approach  has faced  the challenge of performing when given with limited data.

\section{Conclusions and Future Work}

This study first presented  an analyses of RI cases in the  South Pacific and 
South 
Indian Ocean region over the last three decades where it was shown that the 
number   cases does not depend on the duration of the 
cyclone.   The extraction of rapid intensification for 
the detection stage reported a class imbalanced problem that led to high rate 
of false positives. A simple strategy was proposed to include more positive 
cases for detection where the  false positive rate was slightly improved.   
  The limitations of building an efficient system remains due to the  challenges of addressing the class imbalance problem  encountered for  rapid intensification prediction. This motivates further research in using innovative machine learning methods.

  The prediction system could be improved further when more 
data is available. In addition, other factors  such as the sea surface 
temperature, humidity and pressure levels could  be incorporated to check if 
they contribute towards  RI. These could further help in challenges to 
deal with the class imbalanced problem in the detection stage. Although RNNs 
were primarily used as the main model for detection, there is further scope for  other  learning algorithms. Further 
improvements for the detection stage could consider the use of  ensemble 
methods such as bagging and boosting. 

In future work, the proposed system  can be used for  cyclones and hurricanes  
in the rest of the regions such as the Atlantic   Ocean. Machine learning 
paradigms such as transfer learning and multi-task learning could be used for 
improvement as a wide range of cyclone data from different regions are available 
that have distinct features in terms cyclone category, duration and decade or 
year of occurrence.  Real-time implementation can be deployed through cloud 
computing infrastructure for computation and web services  for mobile 
applications for disaster management.

\bibliographystyle{IEEEtran}
\bibliography{merged,cyclone,reference,benchmark}

\begin{thebibliography}{10}
\providecommand{\url}[1]{#1}
\csname url@samestyle\endcsname
\providecommand{\newblock}{\relax}
\providecommand{\bibinfo}[2]{#2}
\providecommand{\BIBentrySTDinterwordspacing}{\spaceskip=0pt\relax}
\providecommand{\BIBentryALTinterwordstretchfactor}{4}
\providecommand{\BIBentryALTinterwordspacing}{\spaceskip=\fontdimen2\font plus
\BIBentryALTinterwordstretchfactor\fontdimen3\font minus
  \fontdimen4\font\relax}
\providecommand{\BIBforeignlanguage}[2]{{%
\expandafter\ifx\csname l@#1\endcsname\relax
\typeout{** WARNING: IEEEtran.bst: No hyphenation pattern has been}%
\typeout{** loaded for the language `#1'. Using the pattern for}%
\typeout{** the default language instead.}%
\else
\language=\csname l@#1\endcsname
\fi
#2}}
\providecommand{\BIBdecl}{\relax}
\BIBdecl

\bibitem{HollidayThompson1979}
C.~R. Holliday and A.~H. Thompson, ``Climatological characteristics of rapidly
  intensifying typhoons,'' \emph{Monthly Weather Review}, vol. 107, pp.
  1022--1034, 1979.

\bibitem{DeMariaEtAl2002}
M.~DeMaria, R.~M. Zehr, J.~P. Kossin, and J.~A. Knaff, ``The use of goes
  imagery in statistical hurricane intensity prediction,'' in \emph{25th
  Conference on Hurricanes and Tropical Meteorology}, San Diego, CA, 2002, pp.
  120 -- 121.

\bibitem{WangZhou2008}
B.~Wang and X.~Zhou, ``Climate variation and prediction of rapid
  intensification in tropical cyclones in the western north pacific,''
  \emph{Meteorology and Atmospheric Physics}, vol.~99, pp. 1 -- 16, 2008.

\bibitem{cloudcycreview}
J.~Houze, R.~A., ``Clouds in tropical cyclones,'' \emph{Mon. Wea. Rev}, vol.
  138, pp. 293--344, 2010.

\bibitem{Gross2001}
J.~M. Gross, ``North atlantic and east pacific track and intensity verification
  for 2000,'' in \emph{55th Interdepartmental Hurricane Conference}, Miami, FL.
  Office of the Federal Coordinator for Meteorological Services and Supporting
  Research, NOAA, B12–B15, 2002, pp. 120 -- 121.

\bibitem{ShayEtAl2000}
L.~K. Shay, G.~J. Goni, and P.~G. Black, ``Effects of a warm oceanic feature on
  hurricane opal,'' \emph{Monthly Weather Review}, vol. 128, pp. 1366--1383,
  2000.

\bibitem{Oropeza2015}
F.~Oropeza and G.~B. Raga, ``Rapid deepening of tropical cyclones in the
  northeastern tropical pacific: The relationship with oceanic eddies,''
  \emph{Atmósfera}, vol.~28, no.~1, pp. 27 -- 42, 2015.

\bibitem{AliCyclone2013}
M.~Ali, D.~Swain, T.~Kashyap, J.~McCreary, and P.~Nagamani, ``Relationship
  between cyclone intensities and sea surface temperature in the tropical
  {Indian Ocean},'' \emph{Geoscience and Remote Sensing Letters, IEEE},
  vol.~10, no.~4, pp. 841--844, July 2013.

\bibitem{Wong2008GACyclone}
K.~Y. Wong, C.~L. Yip, and P.~W. Li, ``Automatic identification of weather
  systems from numerical weather prediction data using genetic algorithm,''
  \emph{Expert Systems with Applications}, vol.~35, no. 1–2, pp. 542 -- 555,
  2008.

\bibitem{Zjavka2016}
L.~Zjavka, ``Numerical weather prediction revisions using the locally trained
  differential polynomial network,'' \emph{Expert Systems with Applications},
  vol.~44, pp. 265 -- 274, 2016.

\bibitem{Wei2012}
C.-C. Wei, ``Wavelet kernel support vector machines forecasting techniques:
  Case study on water-level predictions during typhoons,'' \emph{Expert Systems
  with Applications}, vol.~39, no.~5, pp. 5189 -- 5199, 2012.

\bibitem{Yanchang2006}
Y.~Zhao and S.~Zhang, ``Generalized dimension-reduction framework for
  recent-biased time series analysis,'' \emph{Knowledge and Data Engineering,
  IEEE Transactions on}, vol.~18, no.~2, pp. 231--244, Feb 2006.

\bibitem{Wai-Keung2010}
W.-K. Wong, E.~Bai, and A.~Chu, ``Adaptive time-variant models for
  fuzzy-time-series forecasting,'' \emph{Systems, Man, and Cybernetics, Part B:
  Cybernetics, IEEE Transactions on}, vol.~40, no.~6, pp. 1531--1542, Dec 2010.

\bibitem{MeieEmbeddingAntCol2013}
M.~Shen, W.-N. Chen, J.~Zhang, H.-H. Chung, and O.~Kaynak, ``Optimal selection
  of parameters for nonuniform embedding of chaotic time series using ant
  colony optimization,'' \emph{Cybernetics, IEEE Transactions on}, vol.~43,
  no.~2, pp. 790--802, April 2013.

\bibitem{DeMariaKaplan1994}
M.~DeMaria and J.~Kaplan, ``A statistical hurricane intensity prediction scheme
  (ships) for the atlantic basin,'' \emph{Weather Forecasting}, vol.~9, pp.
  209--220, 1994.

\bibitem{RozoffKossin2011}
C.~M. Rozoff and J.~P. Kossin, ``New probalistic forecast models for the
  prediction of tropical cyclone rapid intensification,'' \emph{Weather and
  Forecasting}, vol.~26, pp. 677 -- 689, 2011.

\bibitem{YuanfeiBPNNCyclone2011}
Y.~Wang, W.~Zhang, and W.~Fu, ``Back propogation(bp)-neural network for
  tropical cyclone track forecast,'' in \emph{Geoinformatics, 2011 19th
  International Conference on}, June 2011, pp. 1--4.

\bibitem{Jaiswal2011Cyclogenesis}
N.~Jaiswal and C.~Kishtawal, ``Prediction of tropical cyclogenesis using
  scatterometer data,'' \emph{Geoscience and Remote Sensing, IEEE Transactions
  on}, vol.~49, no.~12, pp. 4904--4909, Dec 2011.

\bibitem{Elman_1990}
J.~L. Elman, ``Finding structure in time,'' \emph{Cognitive Science}, vol.~14,
  pp. 179--211, 1990.

\bibitem{Werbos_1990}
P.~J. Werbos, ``Backpropagation through time: what it does and how to do it,''
  \emph{Proceedings of the IEEE}, vol.~78, no.~10, pp. 1550--1560, 1990.

\bibitem{Bengio_etal1994}
Y.~Bengio, P.~Simard, and P.~Frasconi, ``Learning long-term dependencies with
  gradient descent is difficult,'' \emph{IEEE Trans. Neural Networks}, vol.~5,
  no.~2, pp. 157--166, 1994.

\bibitem{frasconi_etal1993}
P.~Frasconi, M.~Gori, and A.~Tesi, ``Successes and failures of backpropagation:
  a theoretical investigation,'' in \emph{Progress in Neural Networks. Ablex
  Publishing}.\hskip 1em plus 0.5em minus 0.4em\relax Ablex Publishing, 1993,
  pp. 205--242.

\bibitem{McBrideHolland1987}
``Tropical-cyclone forecasting: a worldwide summary of techniques and
  verification statistics,'' \emph{Bulletin of American Meteorological
  Society}, vol.~68.

\bibitem{Holland2009}
G.~Holland, ``Global guide to tropical cyclone forecasting. bureau of
  meteorology research center, melbourne, australia,''
  \url{http://cawcr.gov.au/publications/BMRC_archive/tcguide/globa_
  guide_intro.htm}, 2009, accessed: January 21, 2015.

\bibitem{RoyKovordanyi2012}
C.~Roy and R.~Kovordányi, ``Tropical cyclone track forecasting techniques - a
  review,'' \emph{Atmospheric Research}, vol. 104-105, pp. 40--69, 2012.

\bibitem{ChandraDayalCEC2015}
R.~Chandra and K.~Dayal, ``Cooperative coevolution of{ Elman }recurrent
  networks for tropical cyclone wind-intensity prediction in the {South
  Pacific} region,'' in \emph{IEEE Congress on Evolutionary Computtaion},
  Sendai, Japan, May 2015, pp. 1784--1791.

\bibitem{ChandraDayalIJCNN2015}
R.~Chandra, K.~Dayal, and N.~Rollings, ``Application of cooperative
  neuro-evolution of {Elman} recurrent networks for a two-dimensional cyclone
  track prediction for the {South Pacific} region,'' in \emph{International
  Joint Conference on Neural Networks {(IJCNN)}}, Killarney, Ireland, July
  2015, pp. 721--728.

\bibitem{ChandraRI2015}
R.~Chandra and K.~S. Dayal, ``Coevolutionary recurrent neural networks for
  prediction of rapid intensification in wind intensity of tropical cyclones in
  the {South Pacific} region,'' in \emph{International Conference on Neural
  Information Processing, {ICONIP} 2015, Istanbul, Turkey, November 9-12, 2015,
  Proceedings, Part {III}}, 2015, pp. 43--52.

\bibitem{Haykin_2008}
S.~Haykin, \emph{Neural Networks and Learning Machines}.\hskip 1em plus 0.5em
  minus 0.4em\relax Prentice Hall, 2008.

\bibitem{KaplanDeMaria2003}
J.~Kaplan and D.~M., ``Large-scale characteristics of rapidly intensifying
  tropical cyclones in the north atlantic basin,'' \emph{Weather Forecasting},
  vol.~18, pp. 1093 -- 1108, 2003.

\bibitem{jtwc}
\BIBentryALTinterwordspacing
(2015) {JTWC} tropical cyclone best track data site. [Online]. Available:
  \url{http://www.usno.navy.mil/NOOC/nmfc-ph/RSS/jtwc/best\_tracks/}
\BIBentrySTDinterwordspacing

\bibitem{Japkowicz2002}
N.~Japkowicz and S.~Stephen, ``The class imbalance problem: A systematic
  study,'' \emph{Intell. Data Anal.}, vol.~6, no.~5, pp. 429--449, 2002.

\bibitem{smartbilo}
\BIBentryALTinterwordspacing
``{Smart Bilo: Computational Intelligence Framework },'' accessed: 02-12-2015.
  [Online]. Available: \url{smartbilo.aicrg.softwarefoundationfiji.org}
\BIBentrySTDinterwordspacing

\end{thebibliography}

\end{document}